\documentclass[journal]{IEEEtran}

\usepackage{xcolor,soul,framed} 

\colorlet{shadecolor}{yellow}
\usepackage[pdftex]{graphicx}
\graphicspath{{../}{../jpeg/}}
\DeclareGraphicsExtensions{.pdf,.jpeg,.png}

\usepackage[cmex10]{amsmath}
\usepackage{array}
\usepackage{mdwmath}
\usepackage{mdwtab}
\usepackage{eqparbox}
\usepackage{url}
\usepackage{hyperref}
\usepackage{amsmath}
\usepackage{amssymb}
\usepackage{ragged2e} 
\usepackage{booktabs,makecell, multirow, tabularx}
\usepackage[ruled,linesnumbered]{algorithm2e}
\usepackage[numbers]{natbib}
\SetKwComment{Comment}{// }{}

\begin{document}
\bstctlcite{IEEEexample:BSTcontrol}
    \title{Interact, Instruct to Improve: A LLM-Driven Parallel Actor-Reasoner Framework for Enhancing Autonomous Vehicle Interactions}
  \author{Shiyu Fang,
        Jiaqi Liu,
        Chengkai Xu,
        Chen Lv,~\IEEEmembership{Senior Member,~IEEE,} \\
        Peng Hang,~\IEEEmembership{Senior Member,~IEEE,} 
        and Jian Sun

\thanks{This work was supported in part by the State Key Lab of Intelligent Transportation System under Project No. 2024-A002, the National Natural Science Foundation of China (52302502, 52472451), the Shanghai Scientific Innovation Foundation (No.23DZ1203400), and the Fundamental Research Funds for the Central Universities.}
\thanks{S. Fang, J. Liu, C. Xu, P. Hang and J. Sun are with the College of Transportation, Tongji University, Shanghai 201804, China, and the State Key Lab of Intelligent Transportation System, Beijing 100088, China. (e-mail: \{2111219, liujiaqi13, 2151162, hangpeng, sunjian\}@tongji.edu.cn).}
\thanks{ C. Lv is with the Nanyang Technological University, 639798, Singapore. (email: lyuchen@ntu.edu.sg). }
}

\maketitle

\begin{abstract}
Autonomous Vehicles (AVs) have entered the commercialization stage, but their limited ability to interact and express intentions still poses challenges in interactions with Human-driven Vehicles (HVs). Recent advances in large language models (LLMs) enable bidirectional human-machine communication, but the conflict between slow inference speed and the need for real-time decision-making challenges practical deployment.
To address these issues, this paper introduces a parallel Actor-Reasoner framework designed to enable explicit bidirectional AV-HV interactions across multiple scenarios. First, by facilitating interactions between the LLM-driven Reasoner and heterogeneous simulated HVs during training, an interaction memory database, referred to as the Actor, is established. Then, by introducing the memory partition module and the two-layer memory retrieval module, the Actor's ability to handle heterogeneous HVs is significantly enhanced. Ablation studies and comparisons with other decision-making methods demonstrate that the proposed Actor-Reasoner framework significantly improves safety and efficiency. Finally, with the combination of the external Human-Machine Interface (eHMI) information derived from Reasoner’s reasoning and the feasible action solutions retrieved from the Actor, the effectiveness of the proposed Actor-Reasoner is confirmed in multi-scenario field interactions.
Our code is available at \url{https://github.com/FanGShiYuu/Actor-Reasoner}.
\end{abstract}


%
\IEEEpeerreviewmaketitle


\section{Introduction}
\IEEEPARstart{A}{s} autonomous driving technology continues to advance, Autonomous Vehicles (AVs) have transitioned from an initial phase focused on technological competition to a new phase centered on commercial deployment. However, when interacting with heterogeneous and uncontrolled Human-driven Vehicles (HVs), AVs often display behavior that appears overly arbitrary or difficult to interpret. This highlights significant deficiencies in their ability to interact effectively and express intentions clearly in complex scenarios.

Various techniques have been proposed to enhance the interaction capabilities of AV, drawing from fields such as artificial potential field theory \citep{LI2020105805, WANG2016306}, game theory \citep{schwarting2019social, camerer2006does}, and cognitive theory \citep{xie2024cognition,zgonnikov2024should}. However, intentions in these approaches often express rely on implicit signals, such as changes in acceleration or lateral movement. While such implicit communication is common and effective among human drivers, it tends to be less effective in AV-HV interaction due to the limited interaction experience and trust that people have with AVs. To address this, some studies have introduced external Human-Machine Interface (eHMI) that use text \citep{bazilinskyy2019survey}, images \citep{zhao2023invisible}, or lights \citep{bindschadel2022active, bindschadel2023using} to explicitly convey the AV intentions. Whereas, existing methods typically depend on fixed-display modes based on predefined rules, which fail to account for the heterogeneity of interaction opponents or bidirectionality of intent communication \citep{9708797, 9629359}. 

Given the remarkable advancements of Large Language Models (LLMs) in natural language understanding and other domains \citep{Radford2019LanguageMA, wei2023chainofthought}, they present a promising avenue for enabling bidirectional intention exchange between AVs and HVs through natural language. Research has shown that LLMs can effectively support various functions in autonomous driving, such as perception\citep{guo2024co, gopalkrishnan2024multi}, decision-making \citep{fu2023drivelikehumanrethinking, cui2024personalized}, and control \citep{sha2023languagempc}. However, current studies often focus on specific scenarios, which limits their generalizability \citep{wen2024dilu}. Moreover, high-level reasoning often requires substantial computation time, and the computational resources of AVs are typically limited, making it difficult to ensure timely responses in dynamic interactions. Therefore, how to achieve real-time response with LLMs also remains to be addressed. Upon deeper analysis, we uncover three main challenges in improving the interaction and intent expression capabilities of AVs: 
\begin{itemize}
    \item Real-time adaptability and interpretability of decision: AVs must rapidly adapt their behavior to dynamic environments while ensuring their actions are interpretable to other road users for safe and efficient interactions.
    \item Heterogeneity and unpredictability of human drivers: AV decisions must be robust enough to handle the diverse driving styles and behaviors of HVs, ensuring effective interactions across heterogeneous HVs.
    \item Complexity and diversity of driving scenarios: A versatile decision framework is needed to navigate diverse traffic environments, accommodating different conflict types and driving rules.
\end{itemize}

\begin{figure}[htbp]
  \begin{center}
  \centerline{\includegraphics[width=3in]{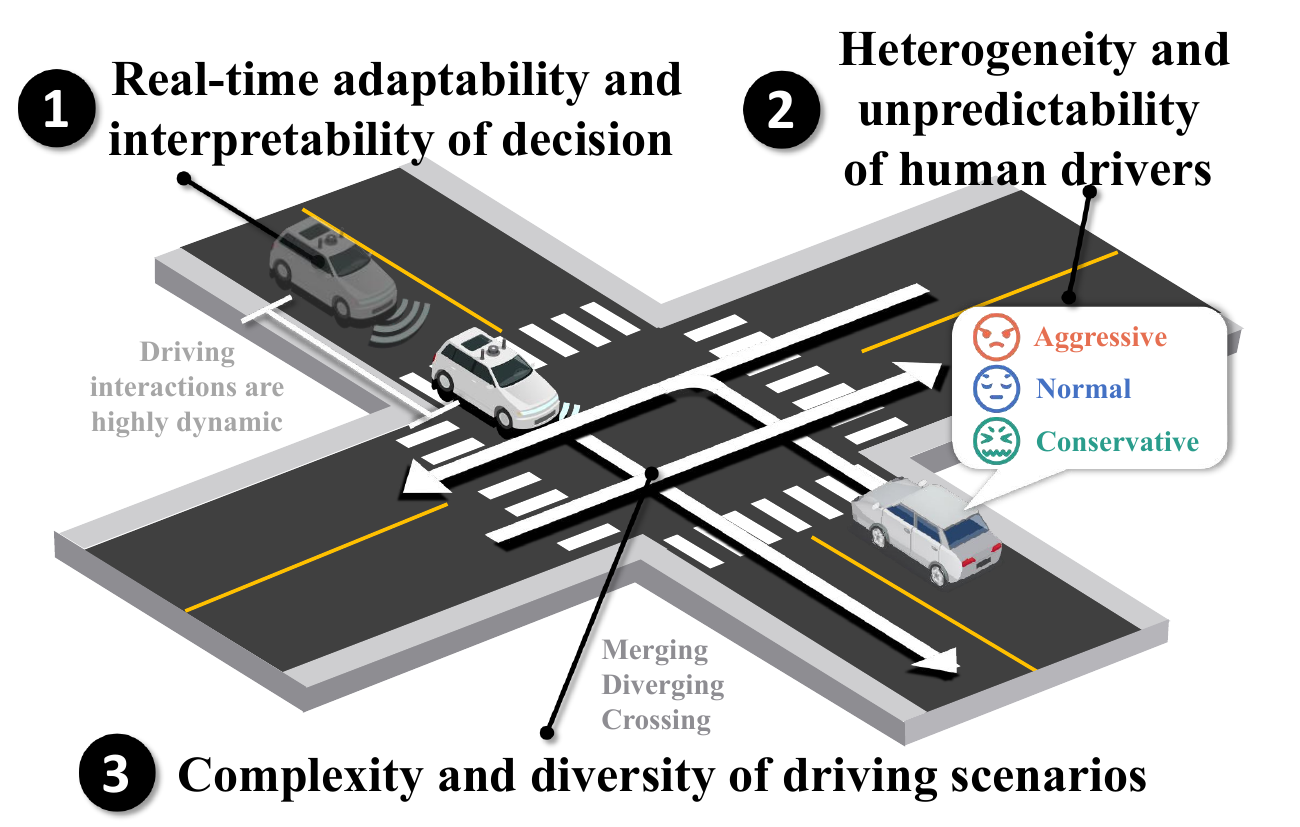}}
  \caption{Main challenges in improving the interaction and intent expression capabilities of AVs.}\label{fig:challenges}
  \end{center}
  \vspace{-0.8cm}
\end{figure}

To address the aforementioned challenges, this paper establishes an Actor-Reasoner framework driven by LLM, aimed at improving the interactive decision-making and intent expression capabilities of AVs. By leveraging interaction memories accumulated through \textbf{Interactions} with heterogeneous simulated vehicles during training and incorporating \textbf{Instructions} from HVs in real-time field testing, the performance of AVs in complex scenarios is significantly \textbf{Improved}. Our contributions can be summarized as follows:
\begin{itemize}
    \item Inspired by the dual-system model of behavioral science theory, which includes "intuitive fast reactions" and "deliberate slow reasoning," this study proposes an Actor-Reasoner framework. By integrating natural language as an intermediary and enabling the parallel operation of the Actor and Reasoner, the framework facilitates explicit bidirectional interaction between HVs and AVs.
    \item The Reasoner combines a standardized definition of interaction scenario states and HV instructions as prompts. By leveraging a localized LLM with Chain-of-Thought (CoT) reasoning to derive HV driving style estimations and eHMI displays, the decision interpretability and generalizations across different scenarios are improved.
    \item The Actor, a segmented interaction memory database formed through interactions between the Reasoner and heterogeneous simulated HVs. It identifies the corresponding memory blocks to be retrieved based on the Reasoner's inference about HV driving styles and retrieves feasible actions from the most similar past memories by combining quantitative scenario descriptions with qualitative experience.
    \item Results from ablation studies and comparisons with other methods show that the Actor-Reasoner effectively enhances safety and efficiency. By applying to multi-vehicle interactions and real-world field tests, demonstrates strong generalizability and practicality. To the best of our knowledge, this is the first study to use LLMs for real-time AV-HV interactions across multiple scenarios.
\end{itemize}

The rest of the paper is organized as follows. Section \uppercase\expandafter{\romannumeral3} presents a literature review of existing studies on improving the interaction and intent expression capabilities of AVs. Section \uppercase\expandafter{\romannumeral3} formulates the problem of our research. Section \uppercase\expandafter{\romannumeral4} introduces the architecture of the proposed Actor-Reasoner framework. In Section \uppercase\expandafter{\romannumeral5}, we validate our framework through several experiments. Finally, conclusions are made in Section \uppercase\expandafter{\romannumeral6}.



\section{Literature Review}
How to enhance the performance of AVs in complex interactive scenarios has garnered significant attention over the past few years. In this section, we summarize the advancements in improving decision-making interactivity, enhancing AV's intention visibility, and exploring the applications of LLMs in autonomous driving to uncover the key challenges that persist in driving interactions and to clarify the motivation behind this study.

\subsection{Interactive Decision-Making}
AVs have now entered the commercial deployment stage both domestically and internationally. While they are capable of handling most driving scenarios, the Beijing Autonomous Vehicle Road Test Report indicates that at least 36\% of takeovers occur during interactions with other road users, indicating that there is still room for improvement in AV decision-making during such interactions \citep{beijing2022}. To enhance the interaction capabilities of AVs, several approaches have been explored, including optimization-based \citep{fang2023real}, rule-based \citep{xing2021toward, kaufman2024effects}, and learning-based \citep{zhou2024reasoning}. \citet{schwarting2019social} proposed method utilizes Social Value Orientation (SVO), which quantifies an agent's level of selfishness or altruism, to improve AV's adaptability in decision-making with heterogeneous objects. \citet{chu2022understanding} applied cognitive rules, such as the Stimulus-Organism-Response (SOR) model, to improve the human-likeness of AV decision-making. \citet{huang2023gameformer} employed a hierarchical game to describe interactions in driving scenarios and further incorporated a layered transformer to capture these interactions, therefore enhancing the AV's interactive capabilities.

Although the above studies effectively improve AV's interaction capabilities by adjusting internal decision-making algorithms, they overlook how to accurately and effectively communicate the AV's intention to other road users after a decision is made. As a result, the behavior exhibited by the AV may lack sufficient interpretability, potentially leading to misunderstandings among other road users.

\subsection{Driving Intent Expression}
According to California’s Department of Motor Vehicles (DMV), 31\% of disengagements are attributed to misinterpretation of intent among interaction participants \citep{DMV2019}. To better convey AV's intention, existing studies focus on both implicit and explicit methods.

Implicit intention expression in AVs can be divided into longitudinal and lateral components based on vehicle dynamics \citep{RETTENMAIER2021103438, yuan2023unified}. \citet{articleliu} analyzed human drivers' implicit intention expression patterns through trajectory and developed an intention-aware acceleration generation method for AVs. \citet{ye2022defining} proposed a progressive lane-change model, where intentions are represented by lateral displacement. Results demonstrated that with multi-step implicit intention expression, lane-change success rate of AV is significantly improved.

Additionally, explicit intention expression can be conveyed through lights, gestures, and honking \citep{lee2021road}. Since lighting changes and honking can provoke other drivers, researchers have explored eHMIs to display intentions currently \citep{dula2003risky, mcgarva2000provoked}. Field experiments by \citet{RETTENMAIER2020175} and \citet{PAPAKOSTOPOULOS202132}. showed that eHMI improves AV intention recognition in ambiguous traffic situations, accelerating interaction convergence. Further studies by \citet{DeyMatviienkoBergerPflegingMartensTerken}. confirmed that combining implicit dynamics with explicit eHMI enables more accurate AV intention recognition by other road users.

However, current research on intention expression primarily focuses on interface designs, with content typically triggered by a fixed set of predefined options \citep{bazilinskyy2019survey}. Such approaches struggle to adapt to interactions in various scenarios and with diverse opponents, resulting in one-way communication. Thus, a bidirectional interaction framework capable of both understanding the intentions of other opponent and effectively expressing its own is still needed.

\subsection{Recent LLM Advancements in AVs}
Recent advancements in LLMs have endowed them with a powerful ability to accurately understand human intentions and express their own ideas effectively, facilitating explicit bidirectional driving interactions. Several studies have already integrated LLMs into various subsystems of autonomous driving \citep{10529537, zhang2024wiseadknowledgeaugmentedendtoend, ge2024llm}. For example, \citet{wen2024dilu} designed a reflection module to help LLMs learn from historical interaction experiences, preventing repeated mistakes. \citet{sha2023languagempc} combined LLMs with Model Predictive Control (MPC) to enable AVs to interpret and reason with high-level information, thereby achieving interpretable and style-variable decisions. \citet{fang2024interactivelearnablecooperativedriving} developed a centralized-distributed negotiation architecture based on LLMs, enabling cooperative decision-making among multiple AVs in mixed traffic. 

However, the slow inference speed of LLMs and the limited computational resources available in AV result in the aforementioned research mostly focused on simulation environments \citep{xu2025telldriveenhancingautonomousdriving}. Though \citet{cui2024personalized, 10491134} introduced Talk2Drive, a groundbreaking implementation that deploys LLMs on real-world autonomous vehicles. Their interaction mainly involves communication between the AV and its passengers, focusing on adjusting the vehicle's driving style based on human instructions after the driving process, rather than facilitating real-time interaction between human drivers and AVs.

In summary, due to the limitations of LLM inference speed, there remains a gap in how to leverage LLMs to enhance the interaction and intention expression capabilities of AVs in real-time field interactions.

\section{Problem Formulation}
\subsection{Assumptions}
To simplify the problem and concentrate on the core aspects of driving interactions between AV and HV, the following assumptions are made:
\begin{itemize}
    \item Positions and speeds are assumed to be perfectly available through communication, without considering errors introduced by perception modules. 
    \item Human drivers are assumed to act consistently with their stated instructions, with no deceptive behaviors.  
    \item Driving intentions are simplified into two distinct categories: yielding or rushing, with the flexibility to switch between these two states.
\end{itemize}

\subsection{Formulation of Driving Interaction}
Driving interaction is characterized by partial observability, stochastic dynamics, and sequential decision dependency, which align closely with the core features of a Partially Observable Markov Decision Process (POMDP). Therefore, we define the POMDP using the tuple \( \mathcal{M}_{\mathcal{G}} = (\mathcal{S}, [\mathcal{O}_i], [\mathcal{A}_i], \mathcal{P}), \mathcal{R}) \) to determine the optimal decision for AV during the driving interaction, where:
\begin{itemize}
    \item \( \mathcal{S} \) denotes the state space, encompassing the states of all agents and the environment,
    \item \( \mathcal{O}_i \) represents the observation space for each agent \( i \in \mathcal{V} \),
    \item \( \mathcal{A}_i \) denotes the action space for agent \( i \),
    \item \( \mathcal{P} \) represents the transition function, capturing the probability of moving from one state to another.
    \item \( \mathcal{R} \) denotes the the immediate reward obtained by performing action \( a \) in state \( S \).
\end{itemize}

At any given time step, each agent \( i \) receives an individual observation \( o_i: \mathcal{S} \to \mathcal{O}_i \) and selects an action \( a_i \in \mathcal{A}_i \) based on a policy \( \pi_i : \mathcal{O}_i \times \mathcal{A}_i \to [0,1] \). The agent then transitions to a new state \( s_i^\prime \) with a probability given by the state transition function \( \mathcal{P}(s^\prime | s, a): \mathcal{S} \times \mathcal{A}_1 \times \cdots \times \mathcal{A}_N \to \mathcal{S} \), where \( N \) is the total number of agents. Finally, to generate the optimal decision during interaction, the problem can be framed as follows:
\begin{equation}
\label{eq:problem}
a_i^* = \arg\max_{a}J(a)= \arg\max_{a} \mathbb{E} \left[ \sum_{t=0}^{\infty} \gamma^t R(a,\mathcal{S}| \pi) \right]
\end{equation}
where \( a_i^* \) is the optimal decision under state \( \mathcal{S} \) with policy \( \pi \), and \( J(a) \) is the abbreviation for the objective function to be optimized.

\section{Methodology}

Complex and dynamic driving interactions place high demands on the real-time performance and interpretability of AV decision-making. Additionally, the heterogeneity of agents and the diversity of scenarios further amplify these challenges. To address these issues, this paper proposes an LLM-driven Actor-Reasoner framework, which combines fast system retrieval with slow system reasoning to enable effective driving interactions tailored to heterogeneous HVs and diverse scenarios.

\subsection{Overall Architecture}
\begin{figure*}[htbp]
  \begin{center}
  \centerline{\includegraphics[width=7in]{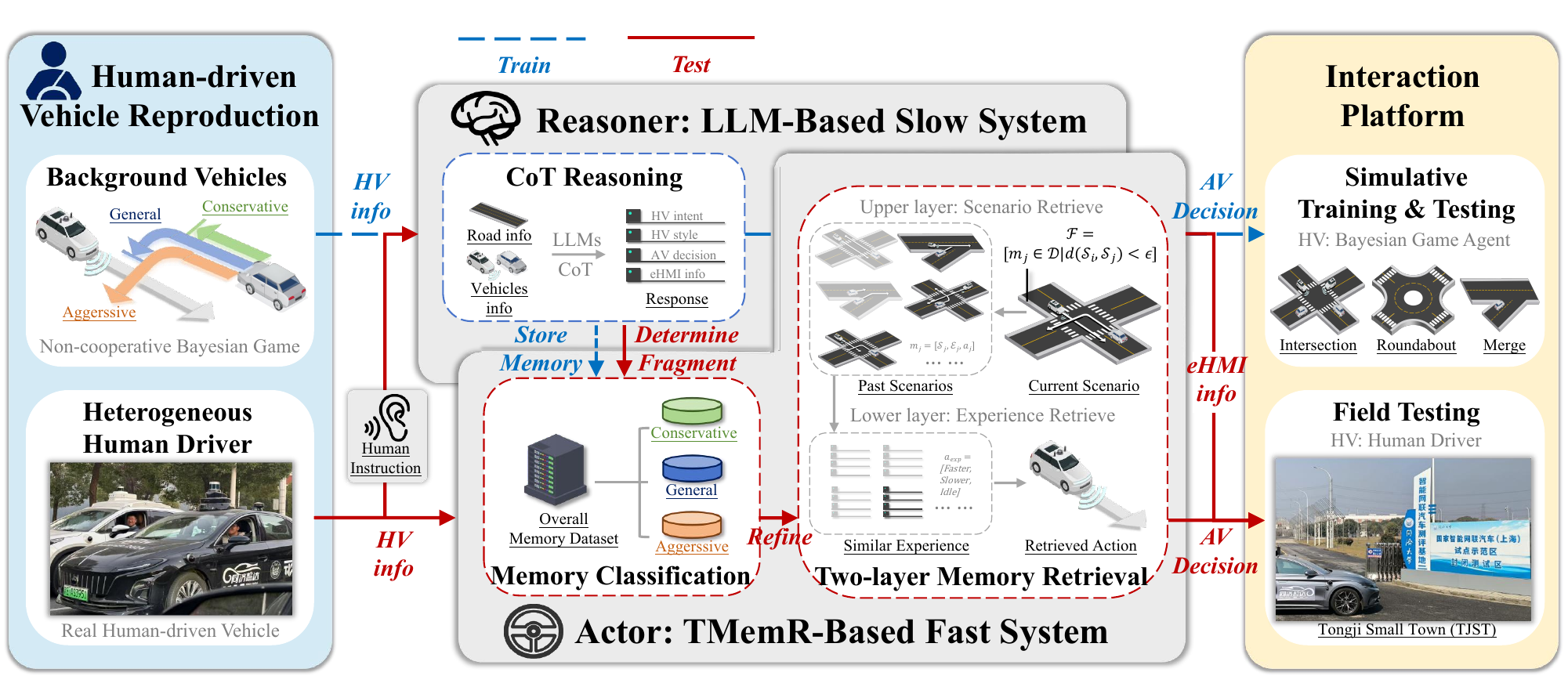}}
  \caption{Overview of the proposed Actor-Reasoner architecture for driving interaction.}\label{fig:framework}
  \end{center}
  \vspace{-0.8cm}
\end{figure*}
Fig.~\ref{fig:framework} depicts the overall framework of the proposed Actor-Reasoner architecture. This architecture draws on two cognitive modes from behavioral science theory when thinking: intuitive fast reactions and deliberate slow reasoning \citep{christakopoulou2024agentsthinkingfastslow, kahneman2011thinking}. That is, in real-world decision-making, the human brain alternates between instinctively generating quick actions based on experience and formulating well-thought-out strategies after careful consideration. Building on this concept, a parallel framework comprising an LLM-based fast system and a Two-layer Memory Retrieval-based (TMemR-based) slow system is designed. This architecture enables the identification of heterogeneous driver styles, the design of eHMI display information, and the rapid generation of AV decisions during driving interactions.

In addition to the Actor-Reasoner architecture, the proposed framework incorporates a heterogeneous driver reproduction module and an environment module. During the training phase, a non-cooperative Bayesian game is utilized to model the decision-making processes of heterogeneous HVs, which are subsequently simulated for interaction within a virtual simulative environment. In the testing phase, experiments are carried out in a real-world test field, where HVs are manually operated by experienced drivers. Further details are provided in Section \uppercase\expandafter{\romannumeral5}.

\subsection{Reasoner}
Language, as a critical medium for bidirectional communication, can serve as a valuable tool for enhancing driving interaction. This subsection introduces the Reasoner, which employs CoT reasoning with LLM to derive the driving style of HVs and eHMI information of AVs.
\begin{figure}[htbp]
  \begin{center}
  \centerline{\includegraphics[width=3.5in]{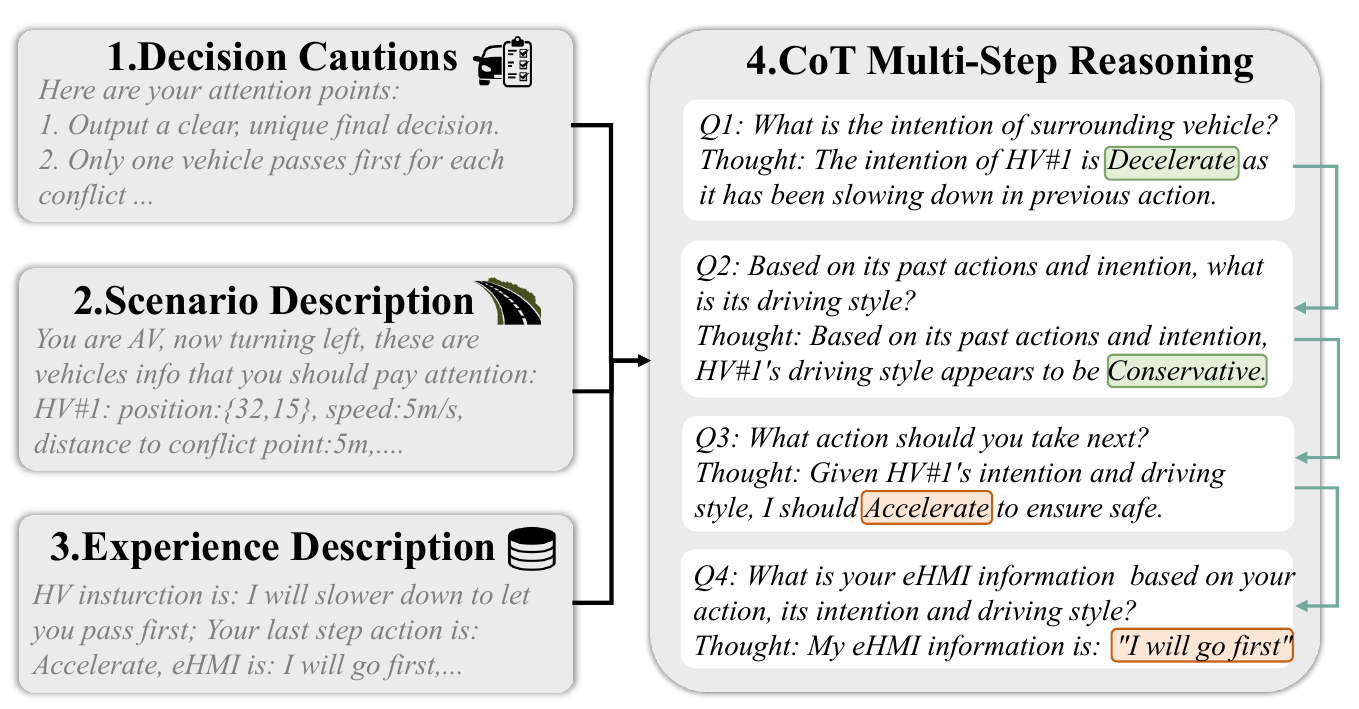}}
  \caption{Illustration of the Reasoner's CoT-based reasoning process.}\label{fig:cot}
  \end{center}
  \vspace{-0.8cm}
\end{figure}

Fig.~\ref{fig:cot} illustrates an example of how the Reasoner employs CoT to complete the reasoning process. First, based on the scenario description and HV instructions, the intent of the interacting HV is estimated. Next, the driver’s style is inferred and categorized as general, aggressive, or conservative. After summarizing the intent and driving style of the HV, the AV's final action is determined. Finally, eHMI information is generated based on this action to share the AV's intention with the interacting HV. This process can be formularized as a recursive problem that follows:
\begin{equation}
\begin{split}
\label{eq:cot}
& a^* = \arg\max_{a}J(a, \arg\max_{\theta}J(\theta, \arg\max_{\tau}J(\tau)))\\
& h^* = \arg\max_{h}J(h,a^*)=\arg\max_{h} \mathbb{E} \left[ \sum_{t=0}^{\infty} \gamma^t R(h,\mathcal{S},\mathcal{I},a^*| \pi_l) \right]
\end{split}
\end{equation}
where \( \tau \) is the estimated HV intention, \( \theta \) is the inferred HV driving style, \( h \) is the displayed eHMI information, \( \mathcal{I} \) is the instruction from HV, and \( \pi_l \) is the pre-trained policy of LLM. By leveraging CoT reasoning to decompose the decision-making task into a series of sub-questions, the reliability and interpretability of the responses of Reasoner in complex driving interactions are improved.

Additionally, the Reasoner plays different roles depending on whether the current interaction is in the training or testing phase. Algorithm.~\ref{train} summarizes the main process in the training process. During training, the designed Reasoner interacts with the simulated heterogeneous vehicles frame by frame, storing the current scenario state and the CoT reasoning results in an external memory database. Ultimately, after filtering out behaviors that lead to inefficient or unsafe driving interactions, a refined memory database is constructed.
\begin{algorithm}[t]
\caption{Train phase of Reasoner.}\label{train}
\KwIn{Vehicle states $\mathcal{S}$}
\KwOut{Overall memory database $\mathcal{D}$}
\Comment{Infer response}
Initialize termination condition $T \leftarrow \text{False}$\;
\While{\textbf{not} $T$}{
Generate prompt based on vehicle states $\mathcal{S}$\;
Feed prompt to Reasoner to generate response\;
Extract HV intention $\tau$, HV driving style $\theta$, AV final action $a$, and eHMI information $\mathcal{h}$ from response as memory $m$\;
\Comment{Store memory}
    \ForEach{$\mathcal{D}_\theta \in \mathcal{D}$}{
        Filter out inefficient and unsafe experience\;
        Add memory to the corresponding memory block $\mathcal{D}_\theta \leftarrow m$\;
    }
    check terminal condition $T$;
}
\end{algorithm}

However, during testing, the reasoning speed of the Reasoner often falls short of meeting the high-frequency and real-time requirements of decision-making in interactions. Therefore, in field tests, the AV’s decisions are rapidly retrieved by the Actor from the memory database. In this phase, the Reasoner is solely responsible for inferring the HV’s driving style and the eHMI displays information to better guide the Actor in generating experienced decisions. 

\subsection{Actor}
Given the high real-time requirements of driving interactions for AV decision-making, this paper proposes a memory retrieval-based decision-making model, termed the Actor. Drawing from the experience accumulated during the training phase, the Actor serves as a lightweight reflection of the Reasoner. Based on the current scenario description and experience description, Actor retrieves the most similar past scenarios and the corresponding decision from the database, thereby rapidly providing a feasible solution. Overall, the Actor consists of two main processes: Memory Partition and Two-layer Memory Retrieval.


\subsubsection{Memory Partition}
In real-world scenarios, human drivers exhibit heterogeneous driving behaviors. To address this variability, during the memory partition phase, the database is partitioned into several blocks based on the driving style of the interacting HV, as inferred by the Reasoner. Each block stores interaction memory specific to a particular driving style. Under default conditions, the Actor retrieves similar memories from the general memory block. Once the Reasoner infers the driving style of the interacting HV, the Actor switches to the corresponding memory block to retrieve a more suitable solution. This adaptive mechanism significantly enhances the AV's ability to interact with HVs of different driving styles, enabling the AV to generate responses that better align with human driver expectations across various scenarios. This memory partition process can be formularized as follows:
\begin{equation}
\mathcal{D} = \bigcup_{\theta=1}^\Theta \mathcal{D}_\theta,\quad \mathcal{D}_\theta = \{ (\mathcal{S}, a) | f(\mathcal{S}) = \theta \} \\
\end{equation}
where \( \mathcal{D} \) is the overall memory database, which stores historical memory when Reasoner interacts with heterogeneous HVs, \( \mathcal{D}_\theta \) represents the $\theta$ memory block, containing interaction data specific to the $\theta$ driving style, and $f(\mathcal{S})$ is the driving style identification function of Reasoner that maps a scenario $\mathcal{S}$ to a driving style category $\theta$.

Memory partitioning effectively narrows the retrieval scope, enabling the system to search only within the memory block associated with the specific driving style. This approach significantly reduces unnecessary data access and improves retrieval efficiency.

\subsubsection{Two-Layer Memory Retrieval}
Once the appropriate memory block is identified, the Actor will further perform a quick search within that memory block to find the most similar memory to the current interaction state and adopt the corresponding AV action from that memory. Each memory consists of three types of information: scenario description $\mathcal{S}_j$, experience description $\mathcal{E}_j$, and corresponding action, as shown in Eq.~\ref{eq:mem}. 
\begin{equation}
\begin{split}
\label{eq:mem}
& m_j = (\mathcal{S}_j, \mathcal{E}_j, a_j) \\
& \mathcal{S}_j = [x_j, y_j, v^x_j, v^y_j, c_j] \\
& \mathcal{E}_j = [\tau_j, \theta_j, \mathcal{I}_j, h_j]
\end{split}
\end{equation}
where $m_j$ represents the smallest basic memory unit in the memory block, the scenario description $\mathcal{S}_j$ includes the positions $[x_j, y_j]$ and velocities $[v^x_j, v^y_j]$ of the vehicles in the longitudinal and lateral directions during the driving interaction, $c_j$ is the conflict information of vehicle $j$, such as Time to Collision (TTC).

The standardized scenario description consists of vehicle state variables such as speed, position, and so on. These numerical values are stored to describe the current interaction situation. The experience description, on the other hand, records the Reasoner’s inference responses, which are textual information for the Actor to invoke. 

Given that memories consist of two types of data, a two-layer memory retrieval method is proposed to search for similar memories more effectively and accurately. First, based on the scenario description, the similarity between the current scenario and the scenarios in the memory block is calculated, and a set of filtered similar scenarios is established. The similarity between the scenarios is measured using the weighted Manhattan distance, which is given by:
\begin{equation}
\begin{split}
\label{eq:scenario retrieve}
&\mathcal{F} = \{ m_j \in \mathcal{D} \mid d(\mathcal{S}_c, \mathcal{S}_j) < \epsilon \} \\
& d(\mathcal{S}_c, \mathcal{S}_j) = \sum_{i=1}^{n} \omega \left| \mathcal{S}_c - \mathcal{S}_j \right|
\end{split}
\end{equation}
where $\mathcal{F}$ is the filtered scenarios that are similar to the current state $\mathcal{S}_c$, $\epsilon$ is the predefined threshold to determine whether states are similar, and $d$ is the distance function that describes similarity with weight $\omega$.

Furthermore, the experience descriptions composed of textual information in similar scenarios filtered by the upper layer are converted into vectors. These vectors are then compared using cosine similarity to ultimately identify the most suitable experienced action. The experience retrieval process can be formularized as:
\begin{equation}
\begin{split}
\label{eq:experience retrieve}
a_{\text{exp}} = a_{j^*},\quad
m_{j^*} = \arg\max_{j \in \mathcal{F}} \frac{\phi(\mathcal{E}_c) \cdot \phi(\mathcal{E}_j)}{\|\phi(\mathcal{E}_c)\| \|\phi(\mathcal{E}_j)\|} 
\end{split}
\end{equation}
where $j^*$ is the index of the most similar experience in filtered memory, and $a_{\text{exp}}$ is the experienced action generated by the Actor. 

Algorithm.~\ref{test} summarizes the main process of the proposed Actor-Reasoner architecture for field test driving interaction. During the interaction, three threads are running in parallel: the Reasoner, the Actor, and the Environment. The Reasoner thread is responsible for inferring the HV driving style for the Actor and generating the corresponding eHMI feedback based on current vehicle states and human instructions. The Actor thread, on the other hand, filters similar scenarios based on the inferred driving style and selects the most similar memory to retrieve the AV action. Finally, the Environment thread updates the scenario and monitors the latest human instructions. This parallel execution structure ensures that the system efficiently handles complex interactions, enabling real-time, context-aware AV decision-making.

\begin{algorithm}[t]
\label{test}
\caption{Test Phase of Actor-Reasoner}\label{algo:test}
\KwIn{Vehicle states $\mathcal{S}$, Overall memory database $\mathcal{D}$}
\KwOut{Final AV action $a^*$, eHMI information $h^*$}

\Comment{Initialization}
\quad Set HV driving style $\theta \leftarrow \text{General}$\;   
\quad Set initial AV action $a \leftarrow \text{IDLE}$\;  
\quad Set termination flag $T \leftarrow \text{False}$\;  
\quad Set other variables $\tau, \mathcal{I}, h \leftarrow \text{None}$\;  
\quad Initialize shared thread lock $l \leftarrow [\mathcal{S}, a, \tau, \theta, \mathcal{I}, h]$\;

\Comment{Start Thread Pool}
\Comment{Parallel Thread 1 (Reasoner):}
\While{not $T$}{
    Retrieve current vehicle state and human instructions from lock $\mathcal{S}_c, \mathcal{I}_c\leftarrow l$\;
    Construct prompt for reasoning\;
    Infer current HV style $\theta_c$ and generate eHMI information $h^*$\ with Reasoner\;
    Update shared thread lock $l \leftarrow \theta_c, h^*$\;
}

\Comment{Parallel Thread 2 (Actor):}
\While{not $T$}{
    Retrieve scenario description from lock $\mathcal{S}_c \leftarrow l$\;
    Generate experience description from lock $\mathcal{E}_c=[\tau_c, \theta_c, \mathcal{I}_c, h_c] \leftarrow l$\;
    \ForEach{$m_j \in \mathcal{D}_{\theta_{c}}$}{
            Filter out similar scenarios $\mathcal{F}$ with Eq.~\ref{eq:scenario retrieve} \;
            Find the most similar memory $m^{*}$ with Eq.~\ref{eq:experience retrieve} \;
            Retrieve the corresponding action $a^*$ from $m^{*}$\;
            Update shared thread lock $l \leftarrow a^*$\;
    }
}

\Comment{Parallel Thread 3 (Environment):}
\While{not $T$}{
    Retrieve current vehicle state and AV action from lock $\mathcal{S}_c,a^* \leftarrow l$\;
    Generate next vehicle state $\mathcal{S}$ based on $\mathcal{S}_c,a^*$ and HV action\;
    Listen for the latest human instruction $\mathcal{I}$\;
    Update shared thread lock $l \leftarrow \mathcal{S}, \mathcal{I}$\;
    Check terminal condition $T$;
}
\end{algorithm}

\section{Experiments and Analysis}
To validate the effectiveness of the proposed Actor-Reasoner architecture, this section is structured into four key parts. First, several state-of-the-art (SOTA) LLMs were compared to identify the foundational model for the Reasoner. Second, ablation experiments were conducted to demonstrate the significant impact of the proposed TMemR mechanism and the incorporation of HV instructions on interaction success rates. Furthermore, the adaptability of the method in multi-vehicle environments is verified, with a comparison of its safety and efficiency against other decision-making approaches. Finally, through field tests with real-time interaction with HVs, the method's applicability to actual AV systems is confirmed.
\subsection{Experiment Settings}
\subsubsection{Reproduction of HV} To reproduce the heterogeneous HVs for evaluating the proposed Actor-Reasoner decision-making framework, drivers were categorized as aggressive, conservative, or normal based on our previous research \citep{fang2024cooperative}. Maximum entropy inverse reinforcement learning was then employed to calibrate their driving preferences. Finally, A non-cooperative Bayesian model was used to simulate the decisions of heterogeneous HVs.

\subsubsection{Interaction Environment} To validate the generalization capability of the proposed method across various scenarios, both simulation training and field tests were conducted in three types of environments: intersections, roundabouts, and merging areas, as shown in Fig.~\ref{fig:framework}. In each case, vehicles were randomly generated at multiple entries. Additionally, during the training phase, only one AV and one HV with a randomly generated driving style were included. 


\subsection{Comparison of the Reasoner’s Foundational Model}
The rapid development of LLMs has led to the release of various models by different companies. However, due to differences in the data used for training and model architecture, their performance in handling different tasks varies. To identify the best-performing model for driving interaction tasks, we compared the success rates and single-step inference times of several models, including Llama3-7B, Qwen2-7B, Mistral-7B, Gemma-7B, and Deepseek-r1-7B, during training. A successful case is defined as one where there are no collisions or deadlocks.

\begin{table}[htbp]
    \caption{Performance of Reasoner with Different LLMs}
    \centering
    \begin{tabular}{c | c | c | c | c }
    \hline
    \multirow{2}{*}{Metrics} & \multicolumn{4}{c}{Models}\\
    \cline{2-5}
     & Llama3 & Qwen2  & Gemma & Deepseek-r1\\
    \hline 
    Success Rate  & \textbf{98\%*} & 82\% & 96\% & 94\% \\
    Inference time (s)  & 2.6 & 3.6 & 1.8 & 10.2 \\
    \hline 
    \end{tabular}
    \label{tab:reasoner llms}
\end{table}

Based on the results in Tab.~\ref{tab:reasoner llms}, Llama3 exhibits the highest success rate in driving interaction and has a moderate inference speed. Therefore, we chose it as the foundational model for the Reasoner to build an interaction memory dataset through interacting with simulated vehicles.

\subsection{Ablation Study of the Actor-Reasoner}
After selecting the Llama3-7B as the foundational model for the Reasoner, ablation experiments were conducted to further compare the effectiveness of the proposed approach. These experiments included four conditions: without instruction, without memory partition, without two-layer memory retrieval, and the complete Actor-Reasoner.

In the absence of HV instructions, HVs do not explicitly communicate their intentions to AVs, limiting interaction to one-way intent transmission via eHMI. When the memory partitioning module is not introduced, AVs assume HVs are homogeneous, performing retrieval only within the general memory block. Lastly, without two-layer memory retrieval, scenario descriptions and experience descriptions were concatenated, and the most similar memories were retrieved using cosine similarity.

\begin{figure}[htbp]
  \begin{center}
  \centerline{\includegraphics[width=3.2in]{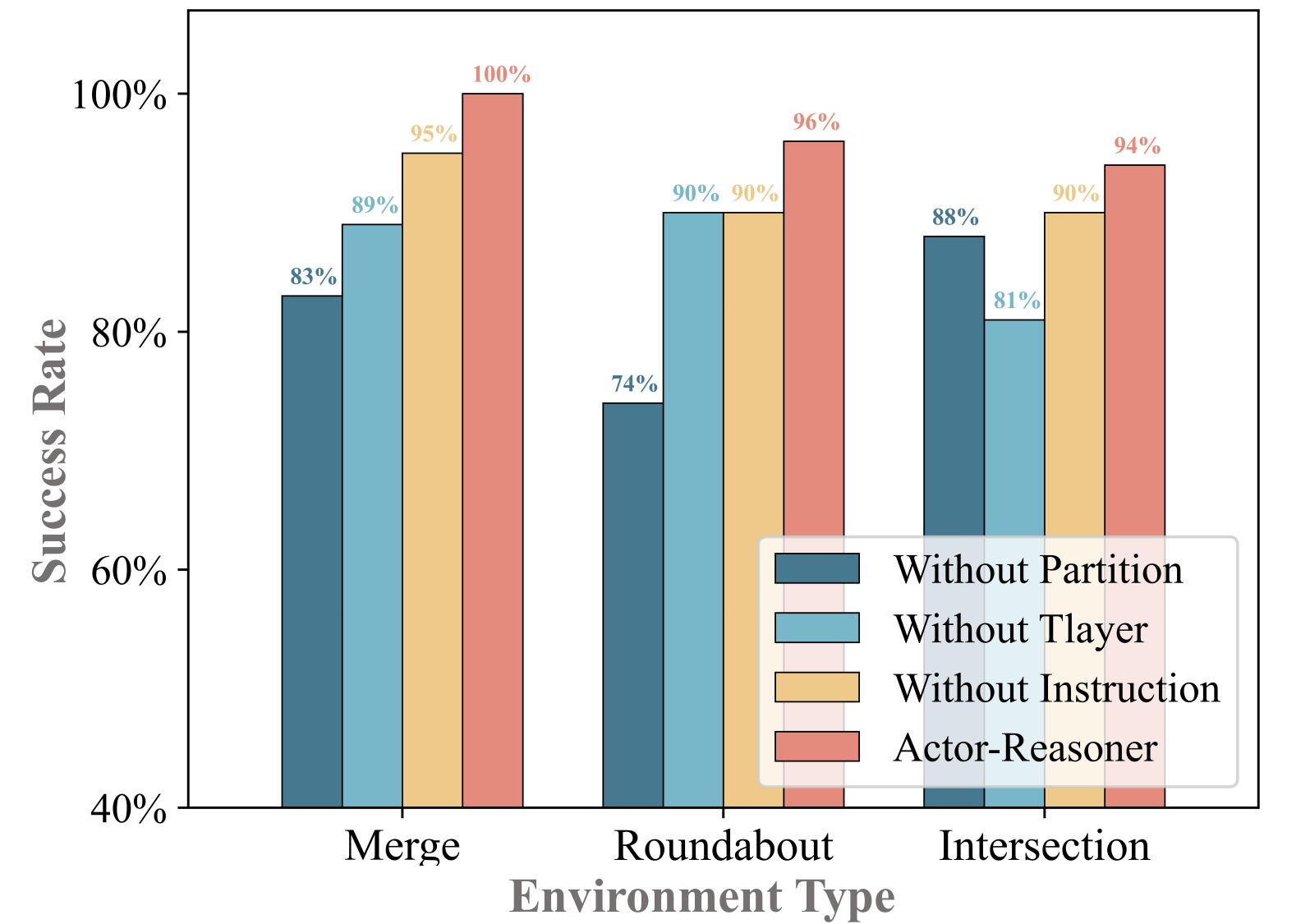}}
  \caption{Ablation Study Results on Success Rates Across Various Scenarios.}\label{fig:ablation}
  \end{center}
  \vspace{-0.8cm}
\end{figure}

Fig.~\ref{fig:ablation} illustrates the model's performance across different scenarios. Overall, in terms of the performance of different model designs, the absence of the memory partition module had the greatest impact on the model, with the success rate dropping by an average of 15\%, and a maximum decrease of 22\% in the roundabout scenario. Additionally, the removal of the two-layer memory retrieval and HV instruction resulted in a 10\% and 5\% decrease in average success rates, respectively. 

On the other hand, from a scenario perspective, the success rate generally decreased from merging areas to roundabouts and intersections. The proposed Actor-Reasoner achieves a 100\% success rate in simpler scenarios like merging areas, where conflicts are relatively straightforward and priority is clear. However, in more complex scenarios, such as intersections, the success rate is lower at 94\%.

\begin{figure}[htbp]
  \begin{center}
  \centerline{\includegraphics[width=3.2in]{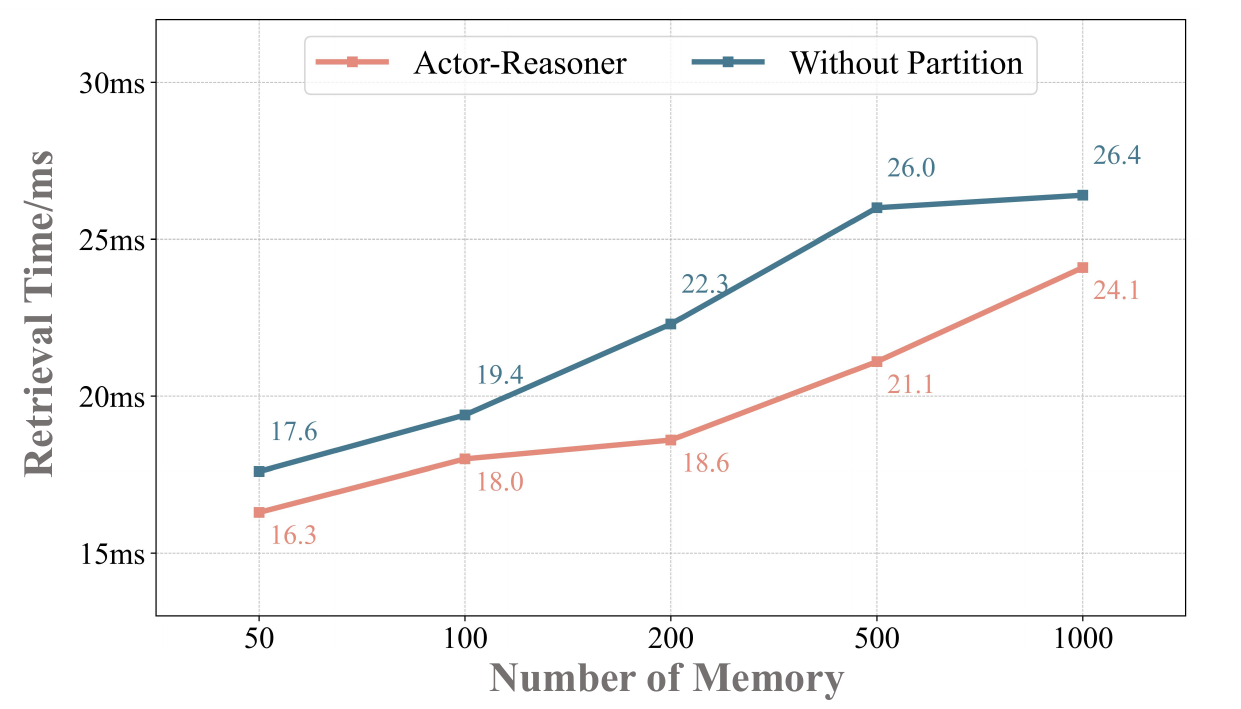}}
  \caption{Retrieval time performance of the Actor with varying numbers of stored memories.}\label{fig:tr_retrieval_time}
  \end{center}
  \vspace{-0.8cm}
\end{figure}

Furthermore, in addition to the improvement in model performance, to validate the significant impact of the memory partition module in enhancing retrieval efficiency, Fig.~\ref{fig:tr_retrieval_time} presents the retrieval time required for the database under different numbers of stored memories. As shown in the figure, the retrieval speed significantly improves after introducing the memory partition module, with an average increase of 12\% and a peak of 19\%. This indicates that the designed memory partition module not only enhances the model's success rate but also significantly boosts decision-making speed, facilitating the application of AVs in field interactions.

\begin{figure*}[htbp]
  \begin{center}
  \centerline{\includegraphics[width=7in]{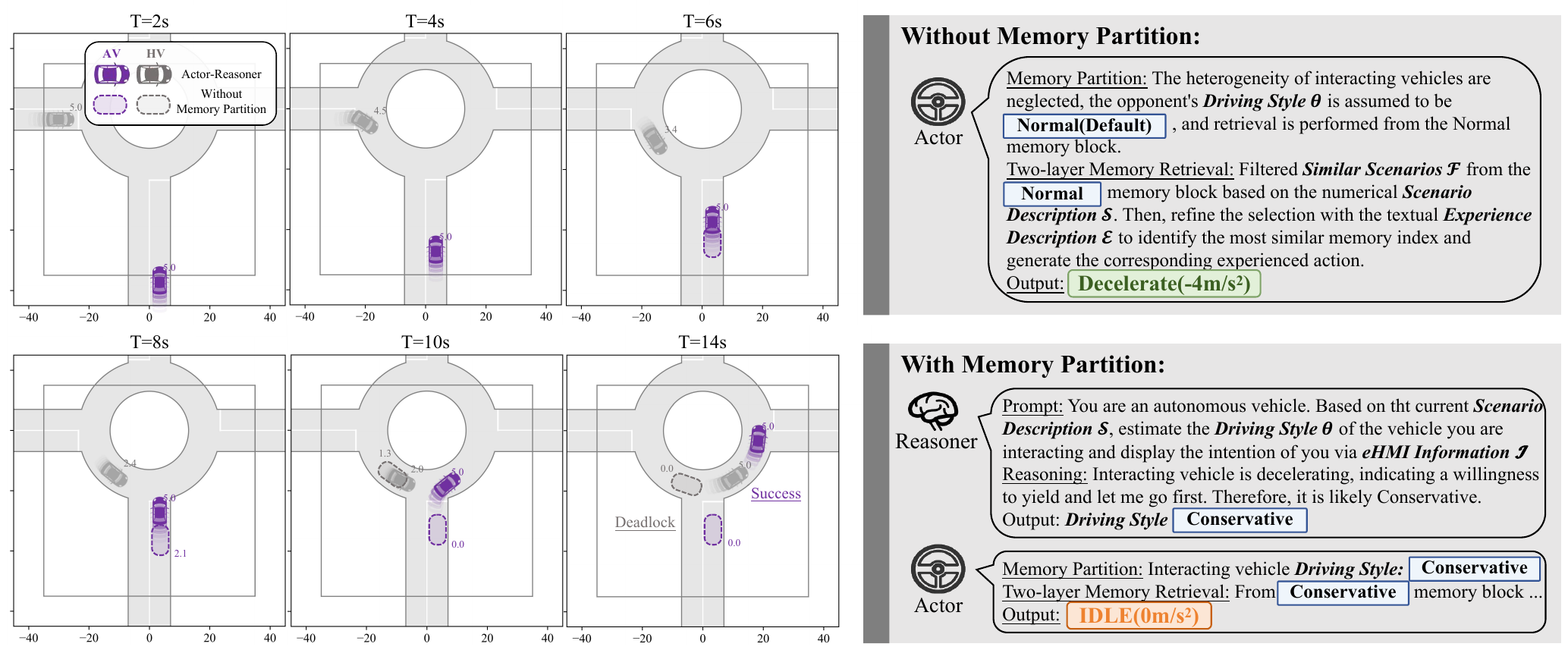}}
  \caption{Comparison of vehicle trajectories and decision-making processes with and without the memory partition module.}\label{fig:partition position}
  \end{center}
  \vspace{-0.8cm}
\end{figure*}

In addition, Fig.~\ref{fig:partition position} compares vehicle trajectories and decision-making processes with and without the memory partition module to visualize the performance differences among models. During the first 4 seconds of interaction, the vehicle's trajectory and decision-making exhibited similar characteristics. However, from the 6s onward, the AV without the memory partition module began to decelerate. This is mainly because the interacting HDVs are already very close to the conflict point, causing the retrieved action suggestions from the entire memory database to be compromised, ultimately leading to a deadlock where both vehicles stop, expecting the other to go first.

In contrast, with the designed memory partition module activated, the Reasoner identified the HV as a conservative type by the 4s and anticipated further yielding. At this point, Actor retrieved experienced action within the conservative memory block. Consequently, the retrieved experience action suggested maintaining the current speed, effectively preventing deadlock.

\subsection{Multi-Vehicle Environment Application and Evaluation}
\begin{figure}[tbp]
  \begin{center}
  \centerline{\includegraphics[width=3.5in]{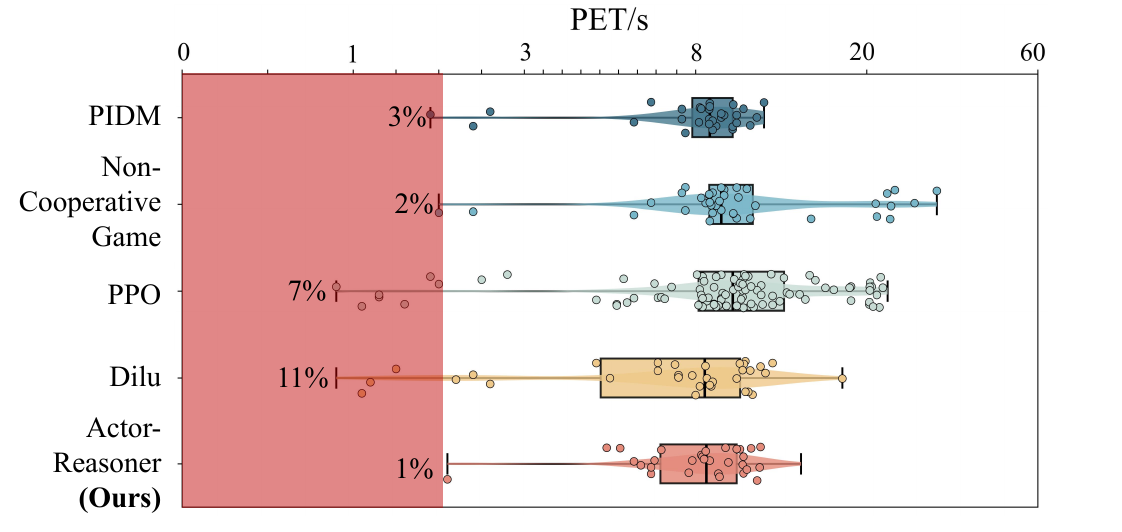}}
  \caption{PET distribution under different decision-making methods.}\label{fig:pet}
  \end{center}
  \vspace{-0.8cm}
\end{figure}
The proposed Actor-Reasoner framework is also applicable in multi-vehicle driving interaction environments. To validate its performance, we selected the intersection scenario and compared the Actor-Reasoner with several other AV decision-making methods, including optimization-based methods (Non-Cooperative Game \citep{10186564}), rule-based methods (P-IDM \citep{deng2025social}), learning-based methods (PPO \citep{9501950}), and LLM-based methods (Dilu \citep{wen2024dilu}) in terms of safety and efficiency. The intersection scenario involves one AV and three HVs with randomly assigned driving styles.

Fig.~\ref{fig:pet} shows the PET distribution of different methods, where the numbers on the left represent the proportion of dangerous interaction events among all interactions. An interaction is considered dangerous if the distance between two vehicles leaving the conflict zone is less than the length of the vehicle. Based on the results from Fig.\ref{fig:pet}, the proposed Actor-Reasoner framework performs the best, with a dangerous interaction rate of only 1\%. Additionally, the rule-based PIDM and optimization-based Non-Cooperative Game performed second, with rates of 3\% and 2\%, respectively. The primary reason is that the two methods respectively rely on spatial planning and constraints to ensure that vehicles do not enter the conflict zone at the same time. Finally, Dilu performed the worst in interactions with an 11\% dangerous rate, which could be attributed to the method originally designed for highway scenarios, where the intensity and patterns of conflicts differ from those in intersections and roundabouts.

\begin{table}[tbp]
    \caption{Efficiency Performance under Different Decision-Making Methods}
    \centering
    \begin{tabular}{c | c | c | c }
    \hline
    \multirow{2}{*}{Methods} & \multicolumn{3}{c}{Travel Velocity (m/s)}\\
    \cline{2-4}
     & Average & Max & Min \\
    \hline 
    PIDM  & 2.9 & 5 & 0.12 \\
    Non-Cooperative Game & 2.5 & 5 & 0.18  \\
    PPO & 3.9 & 4.7 & 2.7  \\
    Dilu & 4.9 & 5 & 4.2  \\
    Actor-Reasoner (Ours) & \textbf{4.0*} & 5 & 0.27  \\
    \hline 
    \end{tabular}
    \label{tab:reasoner llms}
\end{table}

Tab.~\ref{tab:reasoner llms} further compares the differences in travel velocity across methods. Based on our observations of trajectories, we found that Dilu’s average travel speed, which is nearly identical to the speed limit, results from its overly aggressive behavior. In almost all cases, Dilu-driven AVs continuously accelerate until reaching the desired speed, forcing surrounding HVs to brake and avoid collisions. This is the primary reason for its significantly higher rate of dangerous interactions compared to other methods. Therefore, we believe that its travel velocity does not effectively reflect its efficiency, and we excluded Dilu’s performance from the analysis. 

According to Tab.~\ref{tab:reasoner llms}, after excluding Dilu, the proposed Actor-Reasoner achieves the best average velocity at 4 m/s. Additionally, while PPO’s performance in safety is slightly worse than that of the proposed Actor-Reasoner, it can still maintain a minimum travel velocity of 2.7 m/s, indicating its ability to safeguard efficiency. This is an area where our algorithm can be further improved. Finally, PIDM and Non-Cooperative Game performed the worst in terms of efficiency, with average travel velocities of 2.9 m/s and 2.5 m/s, respectively. This low efficiency may stem from their excessive emphasis on safety.

Overall, through the analysis of PET and travel velocity, the proposed Actor-Reasoner demonstrates the ability to maintain high efficiency while ensuring safety, proving its superiority in multi-vehicle interactions.


\begin{figure*}[htbp]
  \begin{center}
  \centerline{\includegraphics[width=7in]{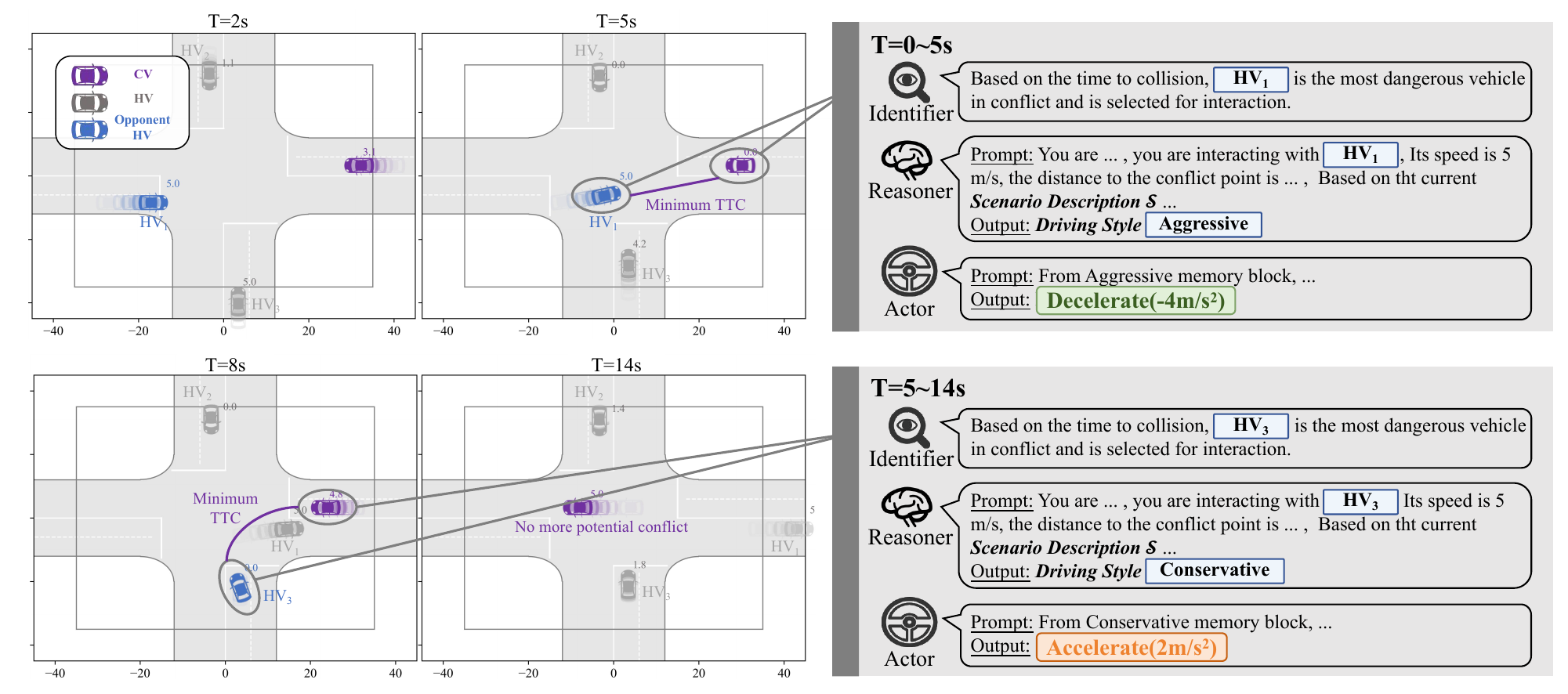}}
  \caption{Vehicle trajectories and decision-making process during multi-vehicle interactions.}\label{fig:multi case}
  \end{center}
  \vspace{-0.8cm}
\end{figure*}

Fig.~\ref{fig:multi case} illustrates the vehicle trajectories and decision-making process during multi-vehicle interactions. Compared to interacting with a single HV, in the presence of multiple potential interacting agents, the AV will first identify the HV with the highest risk based on the TTC from the scenario description and regard it as the interacting opponent. In the first 5s of the interaction, $\text{HV}_{1}$ with a higher speed becomes the primary focus of the AV. The Reasoner recognizes it as an aggressive style, and the Actor retrieves a deceleration action from the corresponding memory block to prevent collision. Once $\text{HV}_{1}$ passes the conflict point, the AV shifts its focus to $\text{HV}_{3}$. Given that $\text{HV}_{3}$ is stationary and shows yielding intent, the Reasoner identifies it as a conservative style and retrieves an acceleration action to avoid deadlock. 

Overall, in multi-vehicle interaction scenarios, the proposed Actor-Reasoner framework effectively recognizes the driving styles of different HVs and retrieves appropriate decisions to prevent deadlock and collisions.

\subsection{Field Testing and Real AV Application}
\begin{figure}[htbp]
  \begin{center}
  \centerline{\includegraphics[width=3.1in]{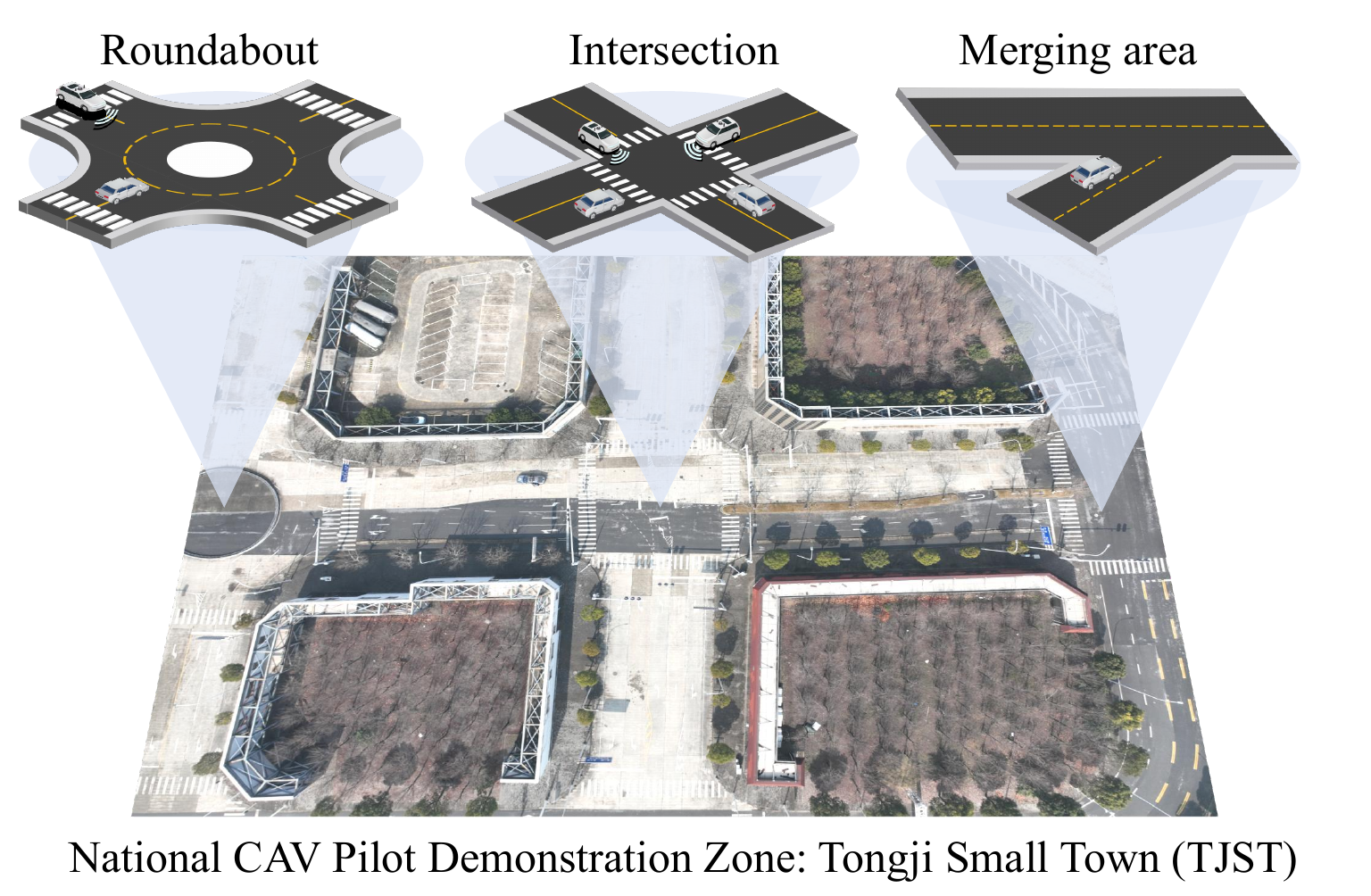}}
  \caption{Field test scenarios of the proposed Actor-Reasoner at Tongji Small Town.}\label{fig:scenario}
  \end{center}
  \vspace{-0.8cm}
\end{figure}

Finally, a field test of the proposed Actor-Reasoner was conducted at Tongji Small Town (TJST) in Shanghai, China, including three scenarios: intersections, roundabouts, and merging areas, as shown in Fig.~\ref{fig:scenario}.

During each interaction, human drivers were asked to express their intentions via voice, which were continuously monitored by the AV’s OBU. The HV instructions were then converted into textual information, serving as the current experience description. Additionally, real-time HV data, such as position information, was transmitted to the AV through vehicle-to-vehicle communication, enabling the formation of the scenario description. Based on the proposed Actor-Reasoner framework, the AV determined the required acceleration and generated corresponding throttle and brake control commands. 

\begin{figure*}[htbp]
  \begin{center}
  \centerline{\includegraphics[width=7in]{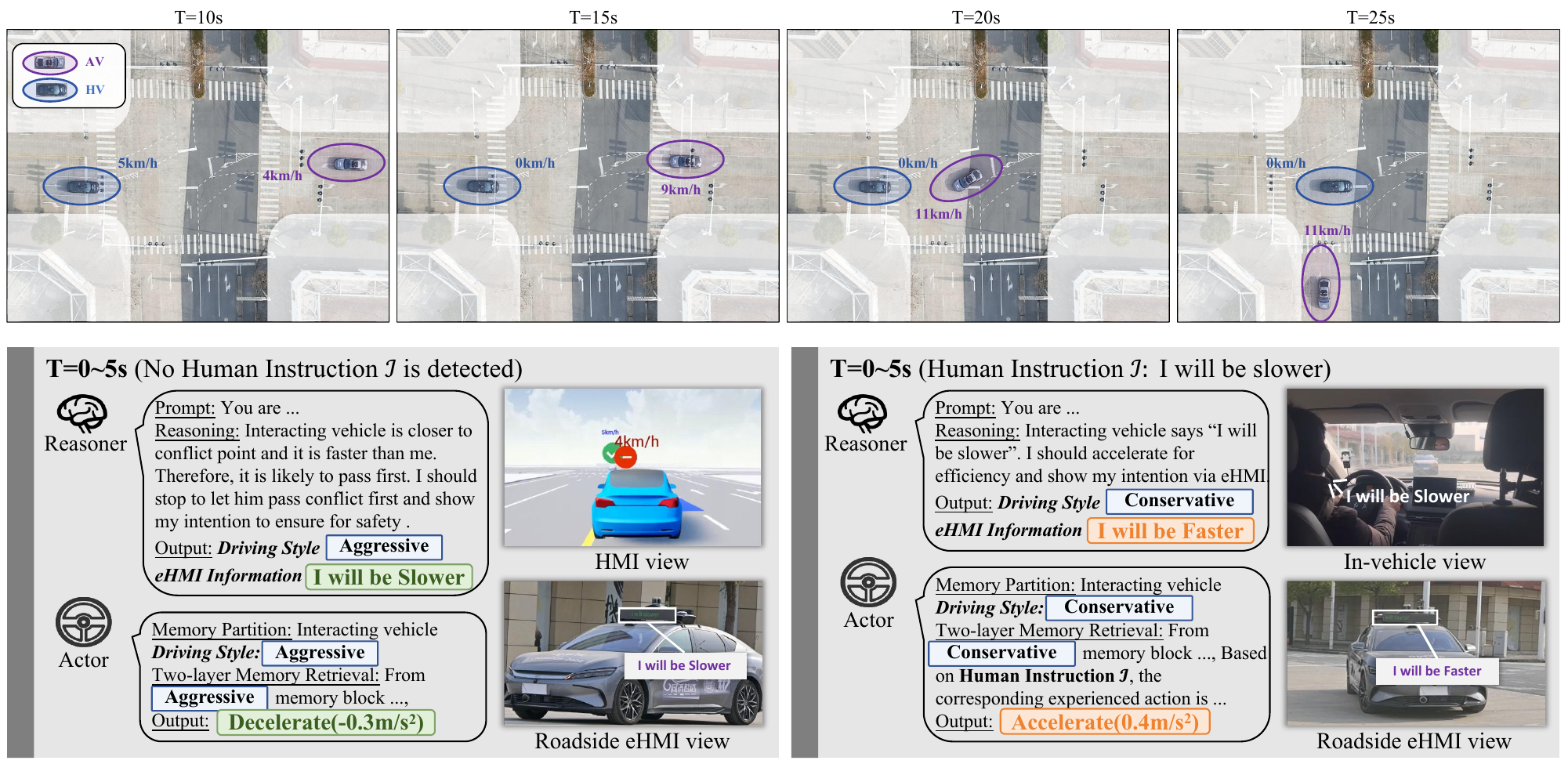}}
  \caption{Vehicle trajectories and decision-making process during field interactions.}\label{fig:field case}
  \end{center}
  \vspace{-0.8cm}
\end{figure*}

To intuitively demonstrate the performance of the Actor-Reasoner framework in real-world interactions, a field test case is shown in Fig.~\ref{fig:field case}. In this scenario, the HV entered the intersection earlier and had a higher initial speed. Consequently, the Reasoner classified the HV as an aggressive driver, prompting the eHMI to display “I will be Slower” to convey the AV’s intention. Based on this assessment, the Actor retrieved a deceleration action to allow the HV to pass through the intersection first.  

However, the HV driver did not follow the AV’s guidance and instead chose to stop at 15s, verbally stating, “I will be slower.” After converting this speech input to text and feeding it into both the Reasoner and the Actor, the Reasoner first interpreted the HV’s intent and, after reasoning, decided to display "I will be Faster" on the eHMI to ensure efficient interaction. The Actor then generated an acceleration decision in real-time, ultimately allowing the AV to pass through the intersection first.

Therefore, through the field test, the proposed Actor-Reasoner framework has been validated in real-world driving interactions for enhancing both AV interaction and intent expression capabilities. Additionally, we visualized the model’s performance in other scenarios, data and videos are available here \footnote{\url{https://fangshiyuu.github.io/Actor-Reasoner/}}.

\section{Conclusion}
The lack of interaction and intent expression capabilities is a key challenge currently faced by AVs on open roads. Recent advancements in LLMs offer a new approach to AV decision-making. To address the conflict between the slow inference speed of LLMs and the high real-time interaction requirements of driving, this paper draws on behavioral science's fast and slow decision-making systems and proposes an Actor-Reasoner framework. By enabling the LLM-based Reasoner to interact with heterogeneous background vehicles during simulation training, an interaction memory, known as the Actor, is constructed. In field tests, with human driver instructions, the Actor and Reasoner operate in parallel to generate AV decisions and eHMI display information, respectively. Through ablation studies and comparisons with other decision-making methods, the Actor-Reasoner framework has demonstrated superior safety and efficiency. Additionally, the method has been applied to multi-vehicle interactions and real-world field tests, proving its strong generalizability and practicality.

However, there is still room for further research in this area. Currently, our study primarily focuses on AV decision-making and intent expression. Given the gradual nature of human cognition, future work will explore how to further optimize the content displayed through eHMI, making it easier for human drivers to accept guidance from AVs. Additionally, we aim to investigate the potential for transferring the proposed framework to the collaborative driving area, extending the focus from optimizing individual behavior to enhancing overall system performance.
\ifCLASSOPTIONcaptionsoff
  \newpage
\fi

\footnotesize
\bibliographystyle{IEEEtranN}
\bibliography{IEEEabrv,Bibliography}

\vfill

\begin{IEEEbiography}[{\includegraphics[width=1in,height=1.25in,clip,keepaspectratio]{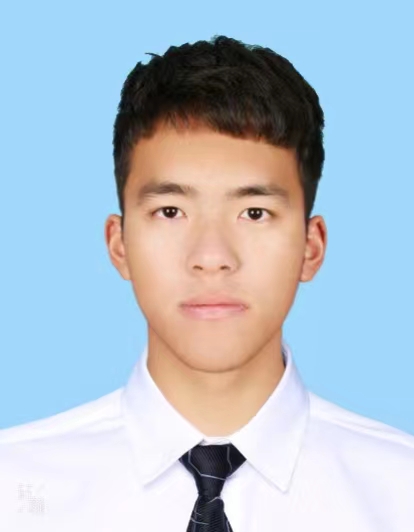}}]{Shiyu Fang}
received the B.S. degree in transportation engineering from Jilin University, Changchun,
China. He is currently pursuing the Ph.D. degree with the Department of Traffic Engineering, Tongji University, Shanghai, China. His main research interests include decision making and motion planning for autonomous vehicles.
\end{IEEEbiography}

\begin{IEEEbiography}[{\includegraphics[width=1in,height=1.25in,clip,keepaspectratio]{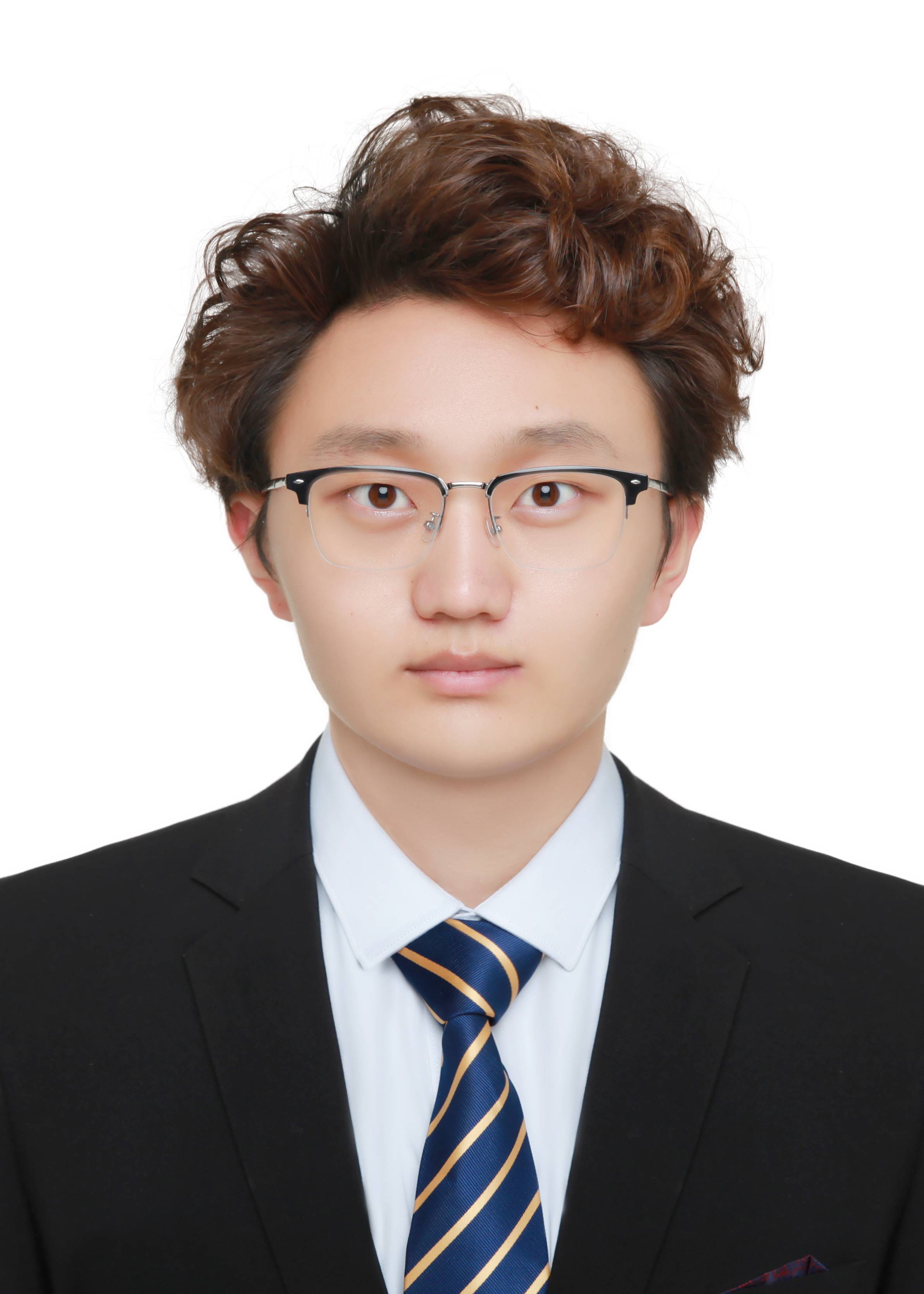}}]{Jiaqi Liu}
received the B.S. degree in transportation engineering from Tongji University, where he is currently pursing the M.S. degree. His research interests include decision-making of autonomous vehicles and data-driven traffic simulation. He won the Best Paper Award of CUMCM in 2020. He is a Visiting Researcher with the Department of Mechanical Engineering, University of California, Berkeley.
\end{IEEEbiography}

\begin{IEEEbiography}[{\includegraphics[width=1in,height=1.25in,clip,keepaspectratio]{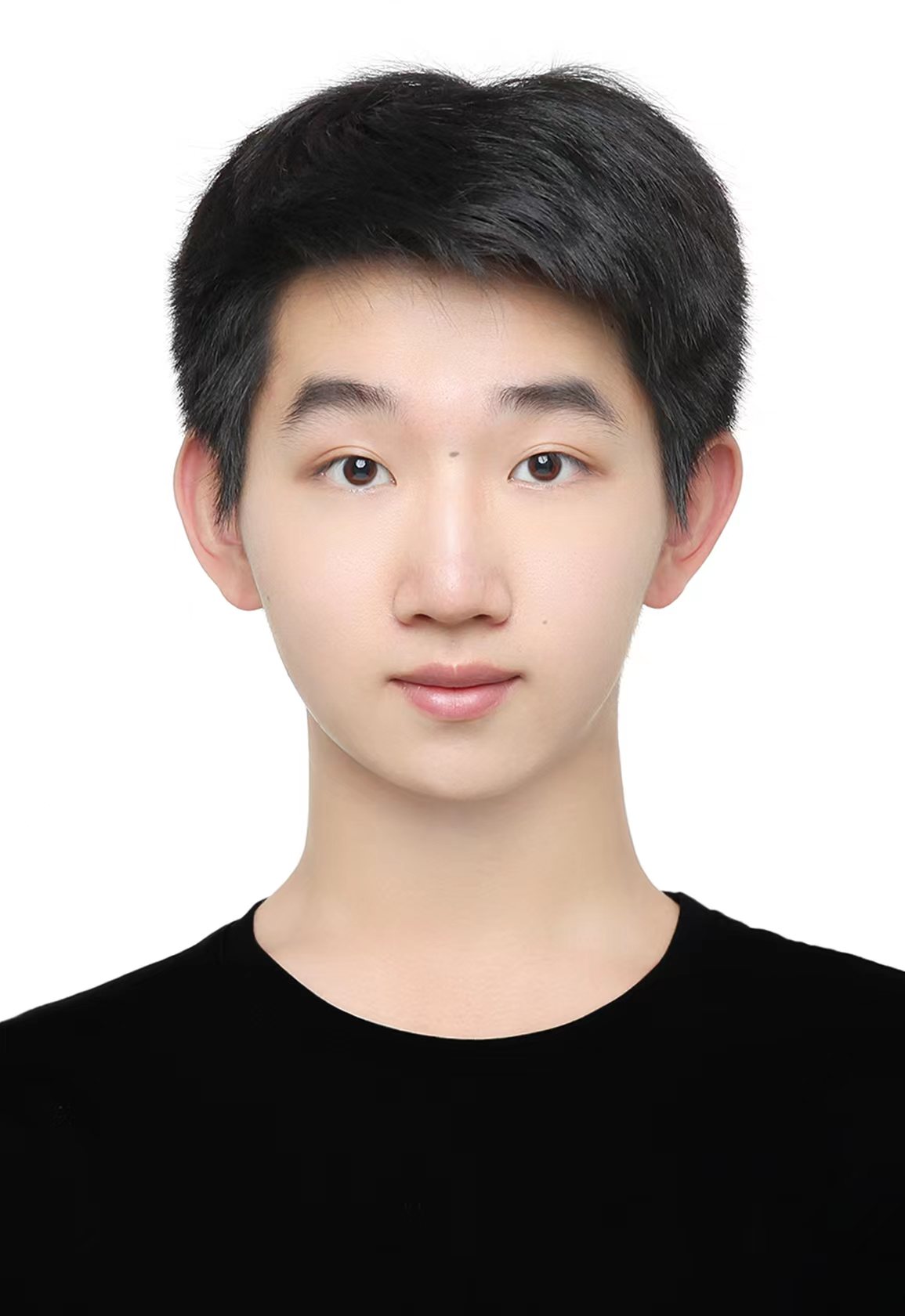}}]{Chengkai Xu}
is an undergraduate student at Tongji University, and will continue his studies at Tongji University as a master's student.
His research interests include decision-making processes, reinforcement learning techniques, and the application of foundation models to enhance autonomous driving systems.
\end{IEEEbiography}

\begin{IEEEbiography}[{\includegraphics[width=1in,height=1.25in,clip,keepaspectratio]{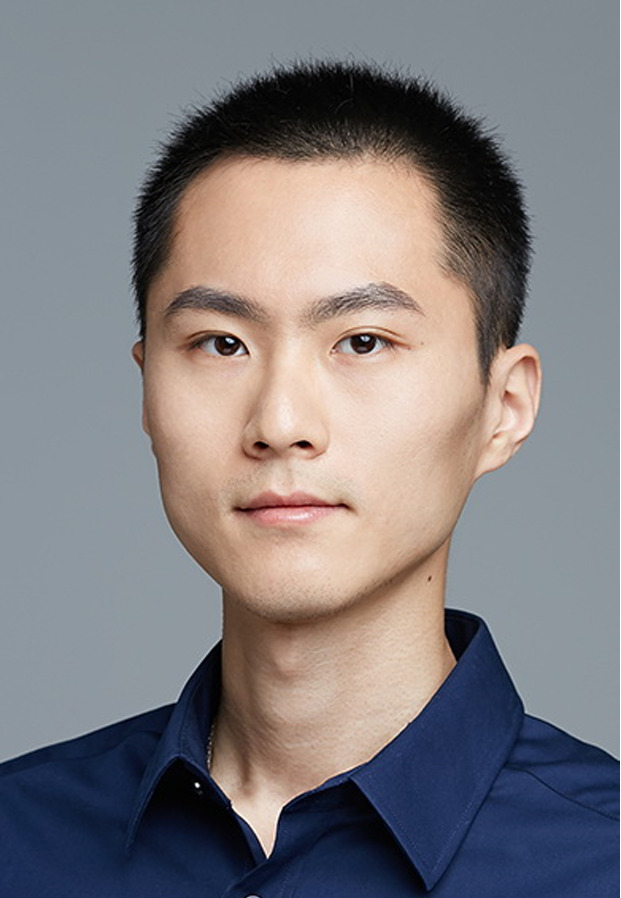}}]
{Chen Lv}  received his Ph.D. degree from the De
partment of Automotive Engineering, Tsinghua
 University, China, in 2016. From 2014 to 2015,
 he was a joint Ph.D. researcher in the EECS
 Dept., University of California, Berkeley.
 He is currently an Associate Professor at
 Nanyang Technology University, Singapore. His
 research focuses on advanced vehicles and
 human-machine systems, where he has con
tributed over 100 papers and obtained 12
 granted patents in China.
 Dr. Lv serves as an Associate Editor for IEEE TITS, IEEE T-VT, IEEE
 T-IV, and a Guest Editor for IEEE ITS Magazine, IEEE-ASME TMECH,
 Applied Energy, etc. He received many awards and honors, selectively
 including the Highly Commended Paper Award of IMechE UK in 2012,
 Japan NSK Outstanding Mechanical Engineering Paper Award in 2014,
 Tsinghua University Outstanding Doctoral Thesis Award in 2016, IEEE
 IV Best Workshop/Special Session Paper Award in 2018, Automotive
 Innovation Best Paper Award in 2020, the winner of Waymo Open
 Dataset Challenges at CVPR 2021, and Machines Young Investigator
 Award in 2022
\end{IEEEbiography}

\begin{IEEEbiography}[{\includegraphics[width=1in,height=1.25in,clip,keepaspectratio]{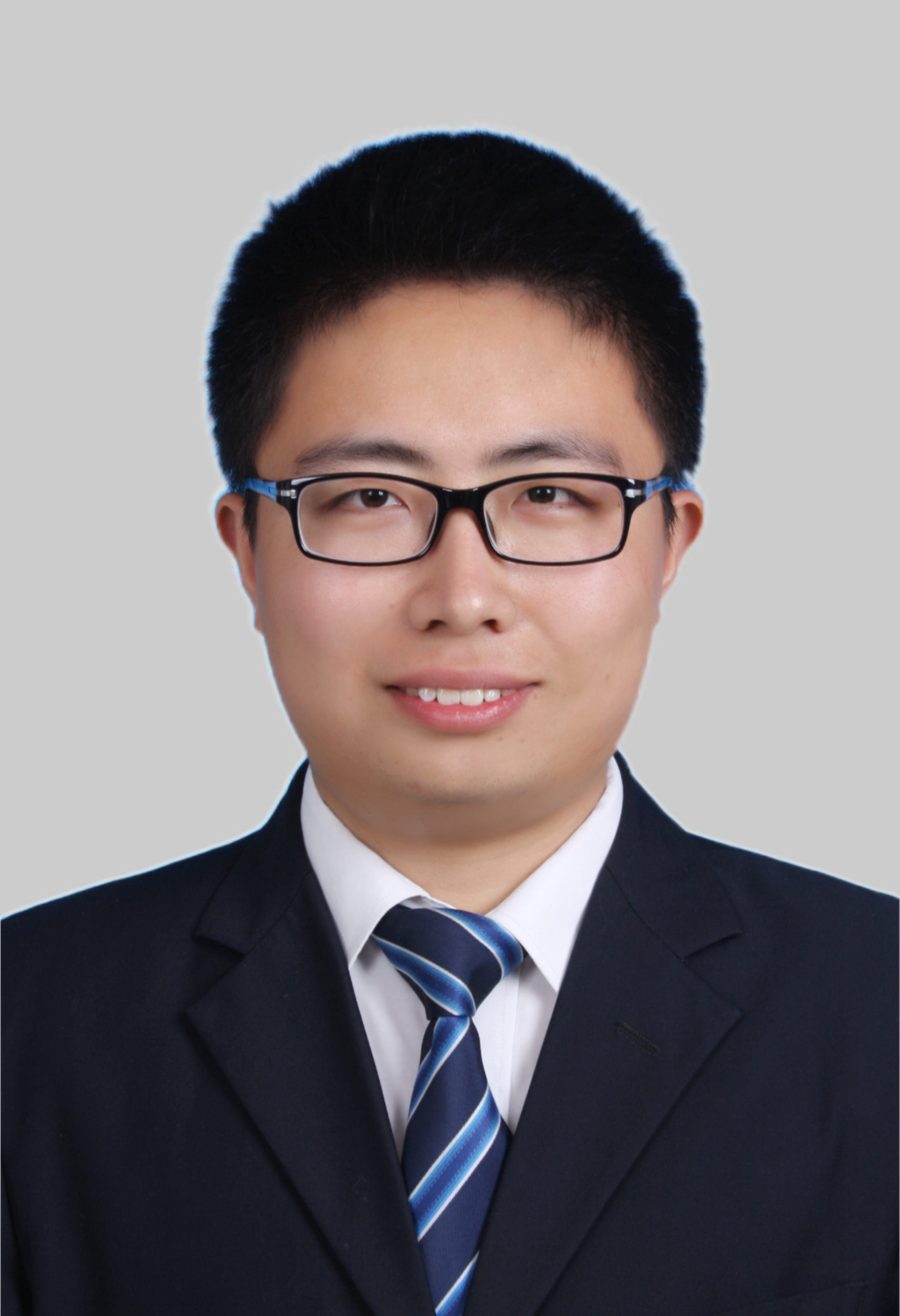}}]
{Peng Hang} is a Research Professor at the Department of Traffic Engineering, Tongji University, Shanghai, China. He received the Ph.D. degree with the School of Automotive Studies, Tongji University, Shanghai, China, in 2019. He was a Visiting Researcher with the Department of Electrical and Computer Engineering, National University of Singapore, Singapore, in 2018. From 2020 to 2022, he served as a Research Fellow with the School of Mechanical and Aerospace Engineering, Nanyang Technological University, Singapore. His research interests include vehicle dynamics and control, decision making, motion planning and motion control for autonomous vehicles. He serves as an Associate Editor of IEEE Transactions on Vehicular Technology, Journal of Field Robotics, IET Smart Cities, and SAE International Journal of Vehicle Dynamics, Stability, and NVH.
\end{IEEEbiography}

\begin{IEEEbiography}[{\includegraphics[width=1in,height=1.25in,clip,keepaspectratio]{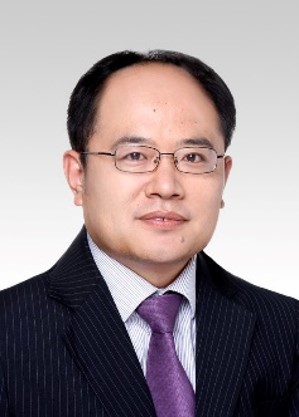}}]
{Jian Sun} received the Ph.D. degree from Tongji University in 2006. Subsequently, he was at Tongji University as a Lecturer, and then promoted to the position as a Professor in 2011, where he is currently a Professor with the College of Transportation Engineering and the Dean of the Department of Traffic Engineering. His main research interests include traffic flow theory, traffic simulation, connected vehicle-infrastructure system, and intelligent transportation systems.
\end{IEEEbiography}

\end{document}